\documentclass[lettersize,journal]{IEEEtran}
\usepackage{amsmath,amsfonts}
\usepackage{algorithmic}
\usepackage{algorithm}
\usepackage{array}
\usepackage{textcomp}
\usepackage{stfloats}
\usepackage{url}
\usepackage{verbatim}
\usepackage{graphicx}
\usepackage{cite}
\usepackage{subfigure}

\usepackage{amsmath}
\usepackage{amssymb}
\usepackage{mathtools}
\usepackage{amsthm}
\usepackage{multicol}
\usepackage{multirow}
\usepackage{microtype}
\usepackage{graphicx}
\usepackage{booktabs} % for professional tables
\usepackage{color}
\newtheorem{theorem}{Theorem}
\newtheorem{proposition}{Proposition}

\usepackage{verbatim}
\theoremstyle{plain}

\theoremstyle{definition}
\newtheorem{definition}[theorem]{Definition}

\theoremstyle{remark}
\newtheorem{remark}{Remark}
\usepackage{amsmath}
\DeclareMathOperator*{\softmax}{softmax}

\begin{document}

\title{Attention Beyond Neighborhoods: Reviving Transformer for Graph Clustering}

\author{Xuanting Xie*, Bingheng Li*, Erlin Pan, Rui Hou, Wenyu Chen, Zhao Kang, ~\IEEEmembership{Member,~IEEE,} \thanks{*These authors contributed equally.}\thanks{Corresponding author: Zhao Kang\\X. Xie, R. Hou, Z. Kang, W. Chen are with the School of Computer Science and Engineering, University of Electronic Science and Technology of China, Chengdu, China; B. Li is with Michigan State University; E. Pan is with Alibaba Group (e-mail: x624361380@outlook.com; libinghe@msu.edu; panerlin.pel@alibaba-inc.com; hours0126@gmail.com; \{cwy, zkang\}@uestc.edu.cn.}}

\markboth{Journal of \LaTeX\ Class Files,~Vol.~14, No.~8, August~2021}%
{Shell \MakeLowercase{\textit{et al.}}: A Sample Article Using IEEEtran.cls for IEEE Journals}


\maketitle

\begin{abstract}
Attention mechanisms have become a cornerstone in modern neural networks, driving breakthroughs across diverse domains. However, their application to graph-structured data, where capturing topological connections is essential, remains underexplored and underperforming compared to Graph Neural Networks (GNNs), particularly in the graph clustering task. GNN tends to overemphasize neighborhood aggregation, leading to a homogenization of node representations. Conversely, Transformer tends to over-globalize, highlighting distant nodes at the expense of meaningful local patterns. This dichotomy raises a key question: Is attention inherently redundant for unsupervised graph learning? To address this, we conduct a comprehensive empirical analysis, uncovering the complementary weaknesses of GNN and Transformer in graph clustering. Motivated by these insights, we propose the Attentive Graph Clustering Network (AGCN)—a novel architecture that reinterprets the notion that ``graph is attention''. AGCN directly embeds the attention mechanism into the graph structure, enabling effective global information extraction while maintaining sensitivity to local topological cues. Our framework incorporates theoretical analysis to contrast AGCN behavior with GNN and Transformer and introduces two innovations: (1) a KV cache mechanism to improve computational efficiency, and (2) a pairwise margin contrastive loss to boost the discriminative capacity of the attention space. Extensive experimental results demonstrate that AGCN outperforms state-of-the-art methods.
\end{abstract}


\section{Introduction}
\IEEEPARstart{G}raph clustering aims to partition nodes into distinct clusters in an unsupervised manner and has been widely applied in real-world scenarios, such as grouping related items in e-commerce graphs or identifying communities of users on social platforms \cite{trivedi2024large}. These methods can be categorized into three types based on their underlying architectures: autoencoder-based methods \cite{AGC,zhu2022collaborative}, contrastive learning methods \cite{shou2025spegcl,SCGC,CCGC}, and shallow methods \cite{CGC,CDC,RGSL}. Currently, all of them leverage Graph Neural Networks (GNNs) \cite{GCN, SGC,chen2025uncertainty}, representative graph learning models, to effectively capture graph structures and node attributes through message-passing.

Attention mechanisms have recently been widely recognized as fundamental components in neural architectures, particularly for modeling long-range dependencies in domains such as natural language processing \cite{devlin2019bert} and speech processing \cite{mehrish2023review}. Variants of the attention framework have also achieved remarkable success in areas like computer vision and code understanding \cite{meng2022adavit,liu2021swin}. However, despite these advancements, the application of attention mechanisms to graph-structured data, where preserving topological relationships is critical, remains relatively underexplored. Specifically, when applied to graph clustering, Transformer-based models \cite{transformer} tend to underperform compared to traditional Graph Neural Networks (GNNs) \cite{GATGC}. This raises a critical question:

\textbf{Is the attention mechanism redundant for graph structures in unsupervised settings?}

In response to this question, we begin by analyzing the respective strengths and limitations of GNNs and Transformers. While GNNs have proven effective for graph clustering due to their ability to model graph structures, they face well-known challenges associated with message-passing, particularly when capturing long-range dependencies. Although increasing the depth of GNNs can theoretically aggregate information from more distant nodes, this often leads to issues such as over-smoothing \cite{li2018deeper} and over-squashing \cite{oonograph}, which hinder the extraction of meaningful graph representations. Despite ongoing efforts to alleviate these problems \cite{chennagphormer}, their inherent limitations remain difficult to overcome completely.

In contrast, Transformer-based models have recently emerged as promising alternatives for node and graph classification tasks due to their ability to capture global information \cite{nguyen2022universal,wu2023sgformer,chennagphormer}. However, they suffer from the ``over-globalization'' problem \cite{xing2024less}: although higher-order neighbors provide additional context, the global attention mechanism tends to overemphasize them, often neglecting more informative lower-order nodes. In unsupervised settings, the absence of ground-truth labels makes it particularly challenging to effectively balance local contributions. Furthermore, the lack of topology-aware message passing in Transformers limits their ability to encode structural information, ultimately hindering their performance in graph clustering.

To better understand these trade-offs in unsupervised settings, we conduct three in-depth empirical studies, which reveal that (1) GNNs tend to excessively exploit the grouping effect within local neighborhoods; (2) incorporating global information can significantly enhance graph clustering performance; and (3) the over-globalization problem is especially detrimental. Having identified these challenges, a natural question arises: \textit{how to apply the attention mechanism to model the graph structures directly in graph clustering?}

Motivated by the ``graph is attention'' concept \cite{dwivedi2021generalization}, we propose a novel Attentive Graph Clustering Network (AGCN). In contrast to existing Graph Transformer approaches that often separate GNN-based message passing from Transformer layers, our method integrates attention mechanisms and structural information into a unified framework. This design places greater emphasis on graph topology than conventional Transformers and effectively alleviates the over-globalization issue, making AGCN particularly suitable for graph clustering. Theoretical analysis against GNN and Transformer supports its effectiveness. Additionally, we incorporate the KV cache technique to reduce computational overhead. We also introduce a novel pairwise margin contrastive loss to enhance the discriminative power of the (Key, Value) space. In summary, our main contributions are as follows.

%By replacing traditional message passing with a purely attention-driven paradigm, AGCN simplifies the architecture while preserving expressiveness. 

\begin{itemize}
    \item{\textit{New perspective.} We propose a new perspective on the ``graph is the attention'' concept, utilizing attention mechanisms to directly model graph structures. Unlike traditional methods, this approach shifts the focus of Transformers toward the graph topology and mitigates the over-globalization issue, making it well-suited for graph clustering.}
    \item{\textit{Novel algorithm.} We propose a novel, fully attention-driven architecture to replace traditional GNNs. Theoretical analysis against GNN and Transformer supports its effectiveness. Furthermore, the complexity is saved through KV Cache idea and the discrimination power of (Key, Value) space is enhanced through a pairwise margin contrastive loss.}
    \item{\textit{SOTA performance.} Extensive experiments on 12 homophilic and heterophilic graph datasets demonstrate the superiority of AGCN.}
\end{itemize}

% The global attention mechanism often overemphasizes higher-order nodes, while valuable information typically resides in lower-order ones.

\section{Related Work}
\subsection{Graph Clustering}
In recent works, several GNN-based methods have been proposed to leverage the structural information in clustering. They can be divided into three kinds \cite{ren2024deep}: (1) Autoencoder-based methods. AGC \cite{AGC} proposes an adaptive high-order GNN method that dynamically selects convolution orders to capture global cluster structures. VGAE-based methods \cite{zhu2022collaborative} integrate pseudo node classification with clustering via variational graph autoencoders, leveraging high-confidence consensus decisions to generate reliable pseudo-labels and enhance representation learning. (2) Contrastive learning methods. DCRN \cite{DCRN} addresses representation collapse in GCNs through a dual-level correlation reduction mechanism that enforces cross-view matrices to approximate identity matrices. SCGC \cite{SCGC} simplifies data augmentation with unshared networks. CCGC \cite{CCGC} leverages high-confidence clustering information into contrastive learning. FPGC \cite{FPGC} proposes to learn a model for each node and a feature cross augmentation in the contrastive framework. (3) Shallow methods. RGSL \cite{RGSL}, CGC \cite{CGC}, MCGC \cite{MCGC}, and FGC \cite{FGC} use a shallow filter to obtain a smooth or sharp representation, which is then input into a graph structure learning framework that incorporates a contrastive learning idea or high-order structure information. However, their reliance on traditional GNNs may lead to excessive use of the grouping effect, leading to a cascade of errors.

\subsection{Graph Transformer}
Transformers \cite{transformer} offer significant benefits in GNNs by utilizing global attention mechanisms, allowing each node to attend to all others in the graph. This helps overcome challenges like over-smoothing and over-squashing, which are common in traditional GNNs. Transformers can capture long-range dependencies and complex relationships with learnable edge weights and fully connected graph structures. Extensive work has achieved remarkable success in graph-level tasks. For example, GPS \cite{gt1} proposes a general and scalable Graph Transformer framework that decouples local message passing from global attention with linear complexity.  UGformer \cite{nguyen2022universal} applies self-attention mechanisms to locally sampled neighborhoods for each node. Motivated by the achievements in graph-level applications, recent works are increasingly investigating how global attention strategies can be applied to node-level tasks. Sgformer \cite{wu2023sgformer} challenges the conventional depth-heavy design of graph Transformers by showing that a single-layer attention is sufficient for effective and scalable representation learning. NAGphormer \cite{chennagphormer} treats each node as a sequence of multi-hop neighborhood tokens via the Hop2Token module. However, these methods suffer from both an oversight of the over-globalization issue and an insufficient exploration of the potential in node clustering tasks.

\section{Preliminaries}
%In this part, we will introduce the relevant notation and background information. 
Denote an undirected graph as $\mathcal{G}=\lbrace \mathcal{V},E,X\rbrace$, where $\mathcal{V} = \lbrace V_1,...,V_N\rbrace$ corresponds to the set of $N$ vertices and $e_{ij}\in E$ represents the connection between vertex $i$ and $j$. The feature matrix with $d$ dimensions is $X=\lbrace X_1,...,X_N\rbrace^{\top}\in \mathbb{R}^{N\times d}$. The adjacency matrix $A \in \mathbb{R}^{N\times N}$ represents the graph structure. The degree matrix $D_{ii}=\sum_j A_{ij}$ is a diagonal matrix where each entry equals the sum of weights connected to node $i$. The normalized adjacency matrix is $\widetilde{A} = D^{-\frac{1}{2}}AD^{-\frac{1}{2}}$, while the normalized $A$ with self-loop is defined as $\widehat{A} = (D+I)^{-\frac{1}{2}}(A+I)(D+I)^{-\frac{1}{2}}$.

\begin{definition}
\textbf{(Grouping effect \cite{GEKDD})} There are two similar nodes $i$ and $j$ in terms of local topology and node features, i.e., $V_i \to V_j \Leftrightarrow \left( \| A_i - A_j \|_2 \to 0 \right) \wedge \left( \| X_i - X_j \|_2 \to 0 \right)$, the matrix $G$ is said to have a grouping effect if $V_i \to V_j \Rightarrow |G_{ip} - G_{jp}| \to 0, \forall 1 \leq p \leq N$.
\end{definition}

In graph learning, the grouping effect occurs naturally as similar nodes converge to closer representations, capturing shared characteristics and improving representation learning. This phenomenon is widely recognized as a key mechanism for enhancing model performance \cite{CDC}.

\begin{figure*}[t]
    \centering
    \subfigure[Cora.]{
        \includegraphics[width=0.2\textwidth]{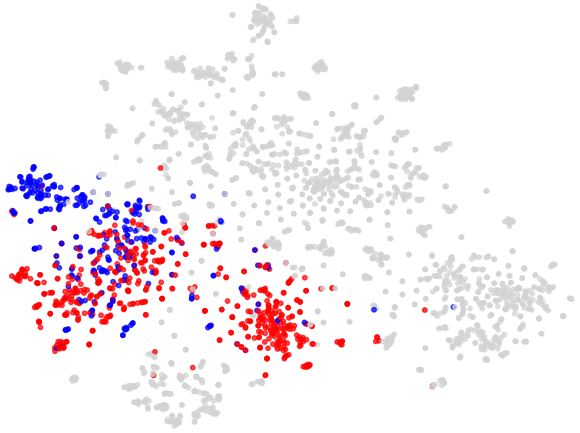}}
        \hspace{0.05\textwidth}
    \subfigure[Pubmed]{
    \includegraphics[width=0.18\textwidth]{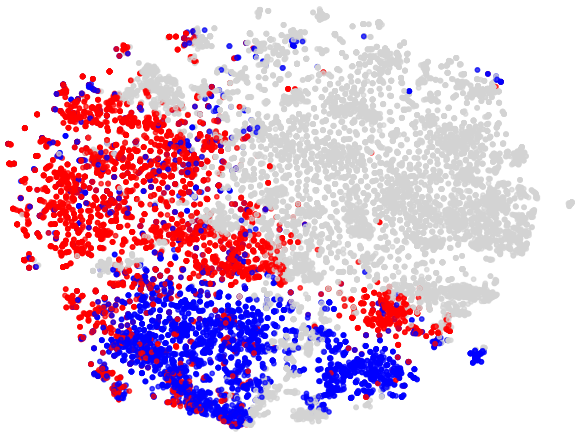}}
            \hspace{0.05\textwidth}
    \subfigure[Cornell]{
    \includegraphics[width=0.2\textwidth]{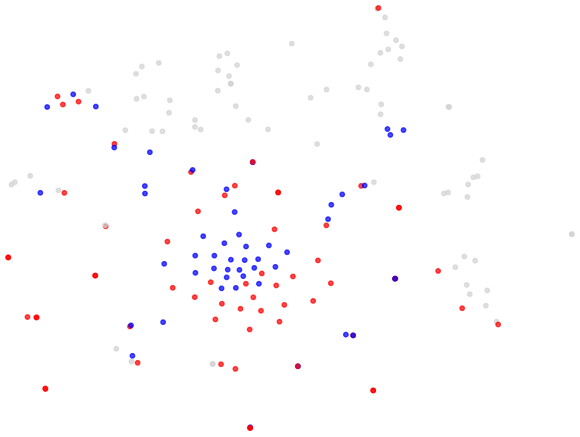}}
            \hspace{0.05\textwidth}
    \subfigure[Chameleon]{
    \includegraphics[width=0.2\textwidth]{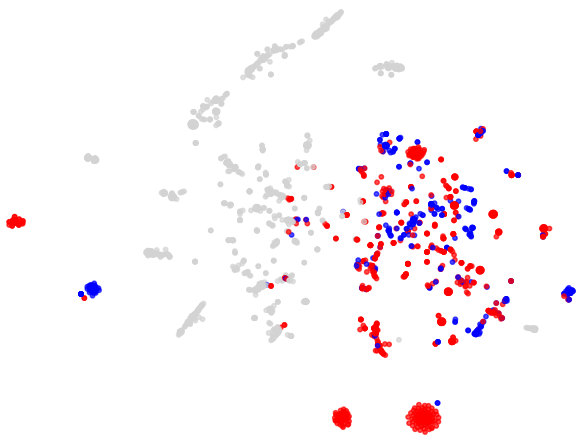}}
    \caption{We apply k-means clustering to the filtered features. Incorrectly clustered nodes are marked in red, while correctly clustered ones are marked in blue. All colored nodes belong to the same cluster.}
    \label{emp1}
\end{figure*}

\begin{figure}[!htbp]
    \centering
        \subfigure[]{
    \includegraphics[width=1.\linewidth]{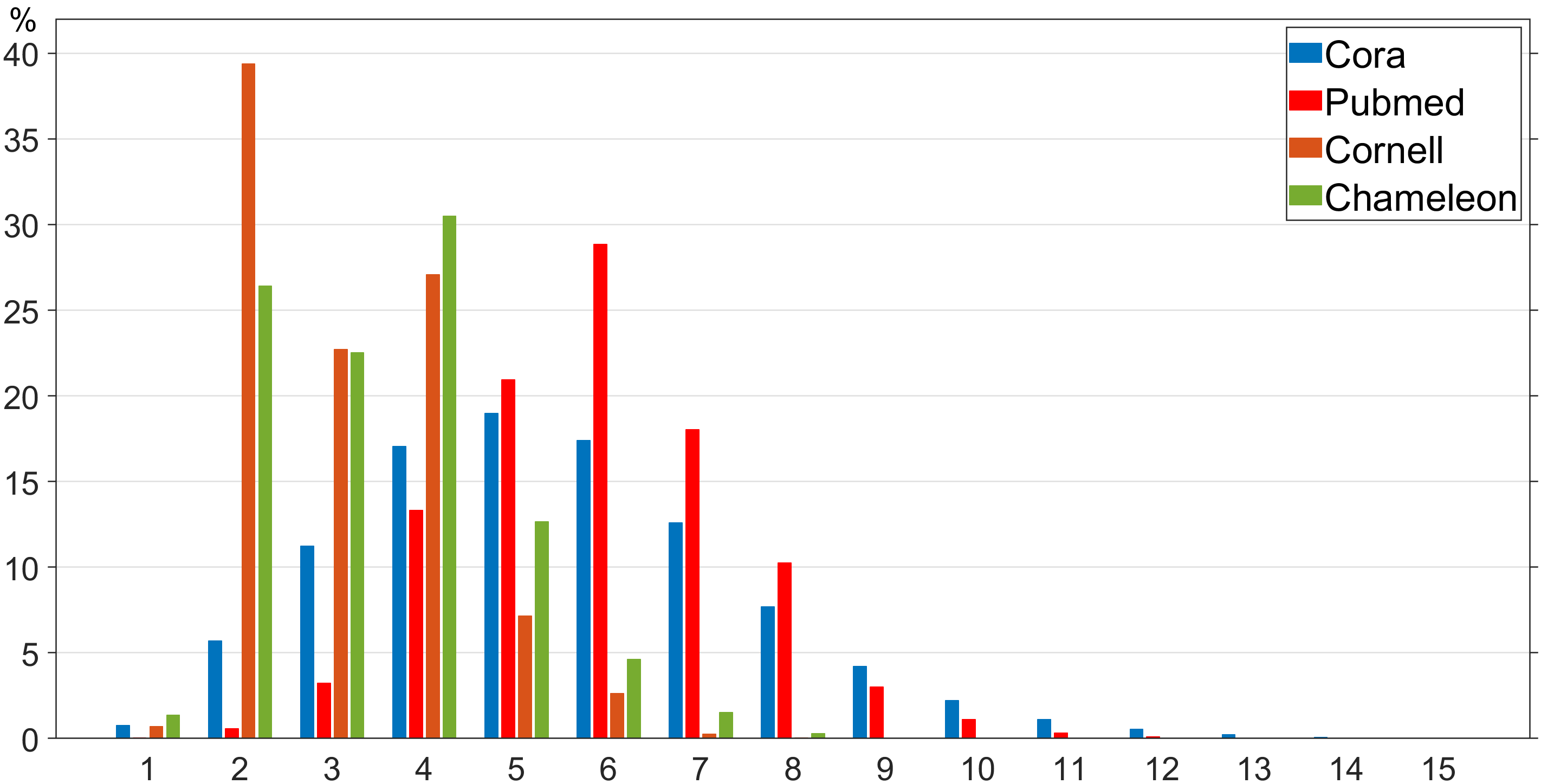}
    }
    \subfigure[]{
    \includegraphics[width=1.\linewidth]{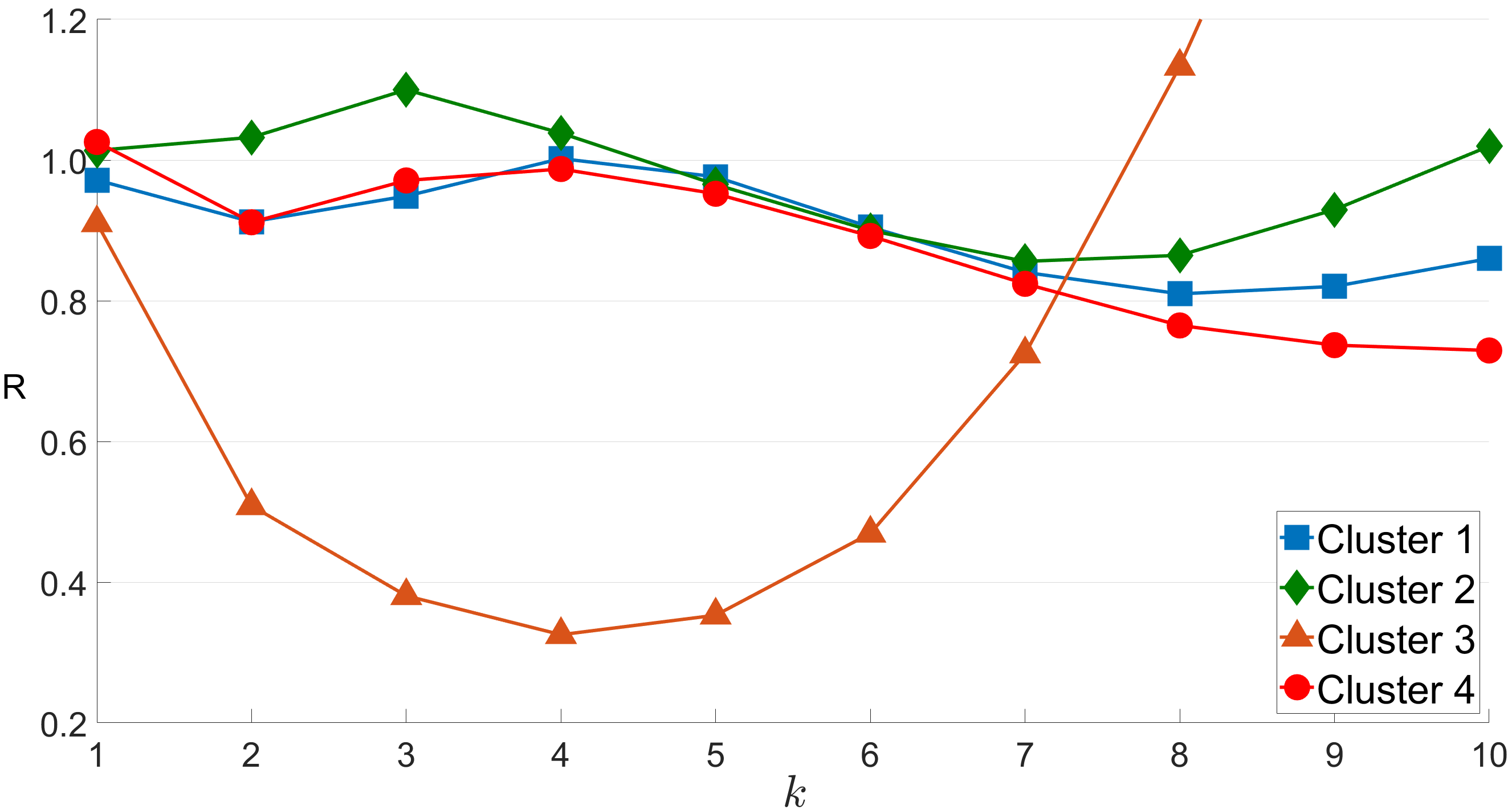}
    }
    \caption{(a) The shortest path distances between nodes within the same cluster on different datasets. (b) SCGC's clustering results on Cora. The relative ratios of their higher-order matrix mean distances compared to the average higher-order matrix distances among all nodes.}
    \label{emp23}
\end{figure}

\section{Empirical Study}
 We use two homophilic graph datasets, Cora and Citeseer, and two heterophilic graph datasets, Cornell and Chameleon. The experiments are conducted directly on the original graphs and features.
 
\textbf{Empirical 1:} In this experiment, we show that classical GNNs are prone to overuse the grouping effect, which can lead to cascading errors and cause similar nodes to be incorrectly clustered. We first apply graph convolution to the features \cite{SGC}, i.e., \(\widehat{A}^kX\), where $k$ is the filter order. We set $k$ to 5, since more than half of the homophilic neighbors can be captured within 5 hops, according to \cite{Global-hete}. The filtered features clearly capture the grouping effect. We then perform K-means clustering and use the t-SNE technique to visualize the clustering results. Specifically, we randomly select one cluster, coloring the correct nodes in blue and the incorrect nodes in red. As shown in Fig. \ref{emp1}, nodes that are close in distance tend to make grouped classification errors, highlighting the need to explore new graph architectures.

\textbf{Empirical 2:} We conduct experiments to highlight the long-range issues. We compute the shortest paths between nodes in the same cluster, as shown in Fig. \ref{emp23} (a). On one hand, we observe that a significant portion of nodes is more than 4 hops apart, highlighting the need to consider long-range dependencies in clustering models. On the other hand, as the distance increases (roughly beyond 10 hops), almost no nodes from the same cluster remain, which can introduce significant noise in the attention mechanism for unsupervised tasks. This aligns with the over-globalization problem.

\textbf{Empirical 3:} We also demonstrate that long-range dependency information can enhance current clustering methods. We apply the well-known contrastive learning method SCGC \cite{SCGC} to obtain the clustering results. Specifically, we follow the parameter settings in the original paper to obtain the clustering results on Cora. Next, we identify the misclustered nodes across different clusters and calculate the relative ratios of their higher-order matrix mean distances compared to the average higher-order matrix distances among all nodes, i.e., $\text{R} = \frac{\sum\limits_{i, u \in c_t}\left\|A_i^k-A_u^k\right\|_2/|c_t|}{\sum\limits_{i, j \in N}\left\|A_i^k-A_j^k\right\|_2/N}$, where \(c_t\) is the misclustered nodes in the $t$-th cluster. We set \(A^k_{ij} \in \{0,1\}\) for a fair comparison. We analyze the results of 4 clusters, as shown in Fig. \ref{emp23} (b). It can be observed that these misclustered nodes exhibit similar local disparities to other nodes (i.e., $k$ = 1), making them difficult to distinguish by traditional GNNs. However, these misclustered nodes are more likely to be grouped based on higher-order information rather than lower-order information (i.e., $k$ = 2-7 in cluster 1 and $k$ = 6-9 in clusters 2-4). Therefore, long-range dependency information is crucial for enhancing traditional clustering methods.

\section{Methodology}
\begin{figure*}[t]
\centering
\includegraphics[width=.8\textwidth]{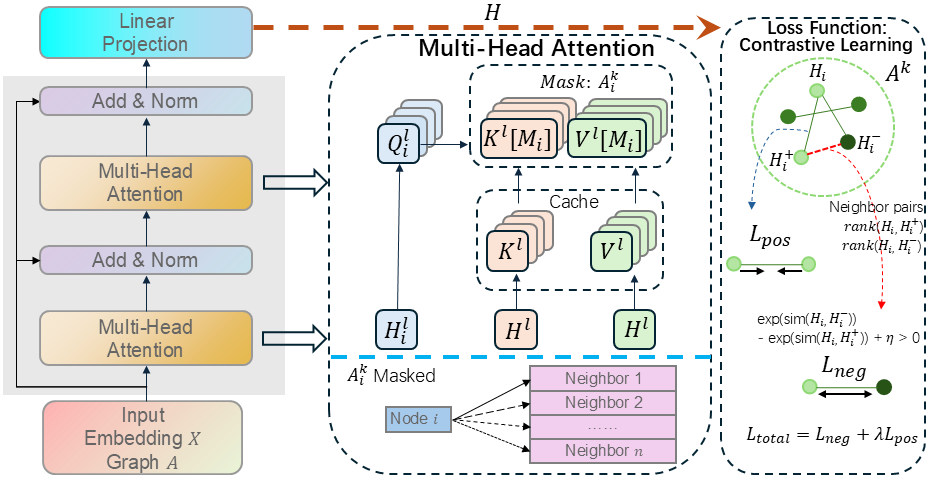} 
\caption{The AGCN framework consists of two main steps: the self-attention architecture to obtain $H$ and the contrastive learning loss for training. AGCN directly models the original structures with an attention mechanism, making the Transformer structure-aware. A pairwise margin loss is introduced to select negative pairs in contrastive learning.}
\label{all}
\end{figure*}

As illustrated in Fig. \ref{all}, our proposed AGCN consists of two modules: the attention module, which leverages an attention mechanism to process the original graph structure, and the contrastive learning module, which enhances discrimination in the (Key, Value) space.
%Motivated by the previous work, 1) in heterophilic networks, vertices with high structural and label similarities are likely far away from each other. 2) On both homophilic and heterophilic graphs, nodes in the same cluster tend to have similar high-order structures. We propose "Graph Attention Transformer Clustering" (AGCN) method. Thus, it's possible that the nodes in the same cluster captures similar attentions.

\textbf{Motivation: Graph is Attention.} Existing approaches often interpret "graph is attention" as using attention to infer or approximate the graph structure, with attention calculated through pairwise dot products. This approach is not applicable in the unsupervised setting due to the over-globalization problem and quadratic complexity. In contrast, we argue that the graph structure inherently encodes a meaningful form of attention, with node connections often reflecting semantic attention. Based on this perspective, we propose directly modeling the graph structure with the Transformer.
% Specifically, the adjacency matrix encodes prior knowledge about which node pairs are likely to interact or influence each other, effectively guiding the message-passing process. Instead of learning attention weights from scratch via pairwise dot products, one can directly leverage the graph structure as a sparse, topology-informed attention mechanism. This not only reduces the computational overhead of learning full attention matrices but also avoids the overuse of the grouping effect.

\textbf{Structure-aware Transformer.} By masking out the attention weights between unconnected nodes, we restrict the attention to follow the topology of the original graph. The global adjacency matrix $A^k$ directly defines the mask token [MASK]. In particular, the mask of the \(i\)-th node \([M_i]\) indicates masking the nodes that are not in $i$'s global neighbors, i.e. \(X[M_i] \in \mathbb{R}^{|\mathcal{N}_i|\times d}\), where \( \mathcal{N}_i \) indicates the neighbors in \(A_i^k\):

\begin{equation}
X^{att}_i=  X[M_i],M_i \in A^k_i.
\end{equation}
 
After the projections, we have \(Q\), \(K\), \(V\) as follows:
\begin{equation}
\begin{aligned}
    &Q_i = X_iW_Q \in \mathbb{R}^{1\times d_Q} ,K_i=X^{att}_iW_K\in \mathbb{R}^{|\mathcal{N}_i|\times d_Q},\\
    &V_i=X^{att}_iW_V\in \mathbb{R}^{|\mathcal{N}_i|\times d_V}
\end{aligned}
\end{equation}
This masking mechanism ensures that the attention space preserves the semantic information encoded by the graph structure.

\textbf{KV Cache.} Due to the high degree of neighbor overlap among different nodes, the resulting projections tend to be highly redundant. Motivated by the KV cache technique \cite{luohekeep}, we propose to store \(K\) and \(V\) in each layer \(l\). Let \(H^{(0)} = X\), we define the \((l+1)\)-th layer's attention as:
\begin{equation}
\begin{aligned}
&Q^{(l+1)} = H^{(l)}W_Q^{(l+1)}, K^{(l+1)} =  H^{(l)}W_K^{(l+1)}, \\
&V^{(l+1)} =  H^{(l)}W_V^{(l+1)}
\end{aligned}
\end{equation}
Then we get the masked attention \(K^{(l+1)}[M_i], V^{(l+1)}[M_i], M_i \in A^k_i\). The cache enables efficient computation by applying a mask to the stored \(K\) and \(V\) associated with each node's high-order neighbors, thereby avoiding redundant projections.

\textbf{Overall Model.} The output of the \((l+1)\)-th layer is:
\begin{equation}
\begin{aligned}
   H^{(l+1)}_i &= softmax(\frac{Q^{(l+1)}_i(K^{(l+1)}[M_i])^T}{\sqrt {d_Q}}) V^{(l+1)}[M_i] \\
   &+ X_iW^{(l+1)}_{res},
\end{aligned}
\end{equation}
where \(W^{(l+1)}_{res}\) is the MLP in the \((l+1)\)-th layer. After the linear projection for the last layer $H^{(l+1)}$, we obtain $H$ as the final output. Besides, to jointly attend to information from different representation subspaces, we apply the multi-head attention mechanism for \(Q, K, V\). The complexity is \(O(Nd_{max})\), where \(d_{max}\) is the maximum degree in \(A^k\). Similar to the vanilla Transformer, it can also be computed through GPUs in parallel.

\textbf{Loss Functions.} To enhance the discriminative power of the (Key, Value) space, we design the training loss with contrastive learning. The connected nodes should have some similarities in most cases \cite{RGSL}, which motivates us to treat edge-connected nodes in $A^k$ as positive samples:
\begin{equation}
\begin{aligned}
&\ell_i=-\log \frac{\sum\limits_{j=1}^{N\setminus \{i\}} \widetilde{A}^k_{ij} \exp \left(\operatorname{sim}\left(H_i, H_j\right) \right)}{\sum\limits_{\mathbf{k}=1}^{N\setminus \{i\}}  \exp \left(\operatorname{sim}\left(H_i, H_\mathbf{k}\right) \right)} \\
&L_{pos} = \frac{1}{N} \sum_{i=1}^N \ell_i,
\label{pos}
\end{aligned}
\end{equation}
where \(\operatorname{sim}\) indicates Cosine Similarity. However, if simply pushing the connected nodes close, the relative difference between edges is ignored. To mitigate this problem, we propose to employ a pairwise margin loss. We first rank all the contrastive examples based on the similarity results in Eq. \ref{pos}. Specifically, for the $i$-th node, we can create a series of example pairs \((H_i^+, H_i^-)\in P_i\), where $P_i$ is the neighbor pairs from $A_i^k$, with $H_i^+$ representing the neighbor with high similarity to node $i$, and $H_i^-$ representing the neighbor with low similarity. For example, if there are $n$ high-order neighbors for node $i$, the $P_i$ has $C_n^2$ elements. These pairs can be used to supervise the model to distinguish between similar and dissimilar nodes in the neighborhood of node $i$. Then, if two neighboring pairs are significantly different, we consider them as negative pairs as follows:
\begin{equation}
\begin{aligned}
    &\ell_i' = \sum\limits_{(H_i^+, H_i^-)\in P_i} L(H_i^+,H_i^-) \\
    &= \sum\limits_{(H_i^+, H_i^-)\in P_i} \max\Bigl\{0, 
        \exp (\operatorname{sim}(H_i,H_i^-)) \\
    &\qquad\qquad - \exp (\operatorname{sim}(H_i,H_i^+)) + \eta \Bigr\} \\
    &L_{neg} = \frac{1}{N} \sum_{i=1}^N \ell_i'.
\end{aligned}
\end{equation}
We set $\eta = \gamma*(rank(H_i^-)-rank(H_i^+))$ following \cite{zhong2020extractive} to reflect the quality difference in these pairs, where $\gamma$ is a parameter controlling the strength and is fixed to 0.0001. $rank(\cdot)$ is based on the similarity rank compared to the node $i$. Our total loss becomes:
\begin{equation}
L_{total} = L_{neg} + \lambda L_{pos}   
\end{equation}
where \(\lambda\) is a balance parameter. The clustering results are obtained by performing K-means on $H$.

%Note that, while both GAT \cite{GAT} and Transformer architectures utilize attention mechanisms, they are fundamentally different. GAT remains a type of GNN that operates under the message-passing paradigm, which suffers from limitations such as over-smoothing and over-squashing, and has difficulty capturing long-range dependencies.

Our method differs from classical GNN-based approaches, like GAT \cite{GAT}, which is attention-based but fundamentally different from Transformers. This novel design offers several advantages. On one hand, it resembles attention-focused methods in NLP \cite{tworkowski2023focused}, focusing more on graph structures rather than irrelevant information, which helps alleviate the over-globalization problem. On the other hand, previous studies \cite{suresh2021breaking,xie2024provable} suggest that nodes with similar neighbors often belong to the same cluster. Specifically, in heterophilic graphs, although nodes from the same cluster may be far apart, they share similar higher-order structures. Therefore, our method performs well on both homophilic and heterophilic graphs.

\section{Experiments}
\subsection{Settings}
We evaluated our approach on 12 datasets, which fall into two categories: homophilic and heterophilic graphs. For heterophilic graphs, we selected 7 datasets: Chameleon, Squirrel \cite{rozemberczki2021multi}, Cornell, Wisconsin, Washington\footnote{http://www.cs.cmu.edu/afs/cs.cmu.edu/project/theo-11/www/wwkb/}, Roman-Empire \cite{Roman}, and the large image relationship network, Flickr \cite{Flickr}. For homophilic graphs, we used 5 datasets: Cora, Citeseer, Pubmed \cite{GCN}, USA Air-Traffic (UAT) \cite{EAT}, and the large social network Twitch-Gamers \cite{twitch}. The homophily ratio is calculated following \cite{Geom-GCN}, with larger values indicating higher homophily. Dataset statistics are provided in Table \ref{tab::datasets}. According to \cite{luan2024heterophily}, Cornell, Wisconsin, and Roman-Empire are especially challenging for GNN models.

To ensure fairness, the experimental settings for each dataset are aligned with those of DGCN \cite{DGCN}, where the optimal parameters are determined through a grid search. The Adam optimizer is used to train the model for 200 epochs until convergence. The number of layers is fixed to 2 and the number of multi-heads is fixed to 4. $d_Q$ and $d_V$ are searched in \{64, 128, 256\}, and the dimensions of the projection layer are set to 100 or 500. $k$ and $\lambda$ are set to \{1,2,...,10\} and \{1e-4,1e-3,...,1e10\}, respectively. All experiments are implemented in the PyTorch platform using an A100-SXM4 80G GPU, 15 workers, Xeon(R) Platinum 8358P CPU and 60G memory. %More implemented details can be found in Appendix \ref{id}.

\begin{table*}[t]
		\centering
	\caption{Statistics information of datasets.}
		\label{tab::datasets}%
    \resizebox{.8\textwidth}{!}{
    \begin{tabular}{ccccccc}
    \toprule
    \multicolumn{2}{c}{Graph Datasets} & Nodes & Dims. & Edges & Clusters & Homophily Ratio \\
    \midrule
    {Heterophilic Graphs} & Cornell & 183   & 1703  & 298   & 5     & 0.1220  \\
          %& Texas & 183   & 1703  & 325   & 5      & 0.0614  \\
          & Wisconsin &251 &1703 &515 &5 &0.1703 \\
          & Washington  & 230   & 1703  & 786   & 5    & 0.1434 \\
          & Chameleon & 2277 & 2325  & 31371 & 5    & 0.2299 \\
          & Squirrel & 5201  & 2089  & 217073 & 5    & 0.2234  \\
          & Roman-empire & 22662  & 300 & 32927 & 18    & 0.0469  \\
          &Flickr &89250 &500 &899756 &7 &0.3195 \\
    \midrule
    {Homophilic Graphs}& Cora  & 2708  & 1433  & 5429  & 7    & 0.8137  \\
          & Citeseer & 3327  & 3703  & 4732  & 6     & 0.7392  \\
          & Pubmed & 19717 &500 &44327 &3 &0.8024 \\
          % & Wiki  & 2405  & 4973  & 16523 & 17    & 0.8024  \\
          & UAT   & 1190  & 239  & 13599 & 4     & 0.6978 \\
          %& AMAP  & 7650  & 745   & 119081 & 8    & 0.8272 \\
          %& EAT   & 399   & 203   & 5994  & 4     & 0.4046 \\
          % & BAT   & 1190  & 239   & 13599 & 4    & 0.4307 \\
          %& Ogbn-arXiv &169,343 &128 &1,166,243 &40 &0.6778 \\
          &Twitch-Gamers &168114 &7 &67997557 &2 &0.5453 \\
    \bottomrule
    \end{tabular}}%
	\end{table*}
    
\subsection{Baselines}
To demonstrate the effectiveness of AGCN, we conduct comparisons against 22 baseline models categorized into 5 groups: 1) Traditional GNN-based approaches, including DAEGC \cite{DAEGC}, MSGA \cite{MSGA}, SSGC \cite{S2GC}, GMM \cite{GMM}, RWR \cite{RWR}, ARVGA \cite{ARVGA}, and AGE \cite{AGE}; 2) Contrastive learning-based methods, such as MVGRL \cite{MVGRL}, SDCN \cite{SDCN}, DFCN \cite{DFCN}, DCRN \cite{DCRN}, SCGC \cite{SCGC}, CCGC \cite{CCGC}, and DyFSS \cite{DyFSS} which incorporate novel data augmentation strategies to improve representation quality; 3) Shallow methods to model the graph structure, including MCGC \cite{MCGC}, FGC \cite{FGC}, and RGSL \cite{RGSL}, where RGSL specifically introduces an innovative $\alpha$-norm to address the heterophily in graph structure learning; 4) Adaptive graph filters to unify homophily and heterophily, such as SELENE \cite{SELENE}, CGC \cite{CGC}, and DGCN \cite{DGCN}. SELENE separates r-ego networks based on node features and structural properties through a dual-channel embedding scheme. CGC adaptively blends high- and low-pass filters to obtain a favorable feature for graph structure learning. DGCN constructs distinct homophilic and heterophilic graphs to better-performing adaptive filters. 5) Transformer-based methods, including GTAGC \cite{GATGC} and TGAE \cite{TGAE}. They directly apply the Transformer in the graph clustering model.

%\vspace{90pt}
\begin{figure}
\centering
\includegraphics[width=0.85\linewidth]{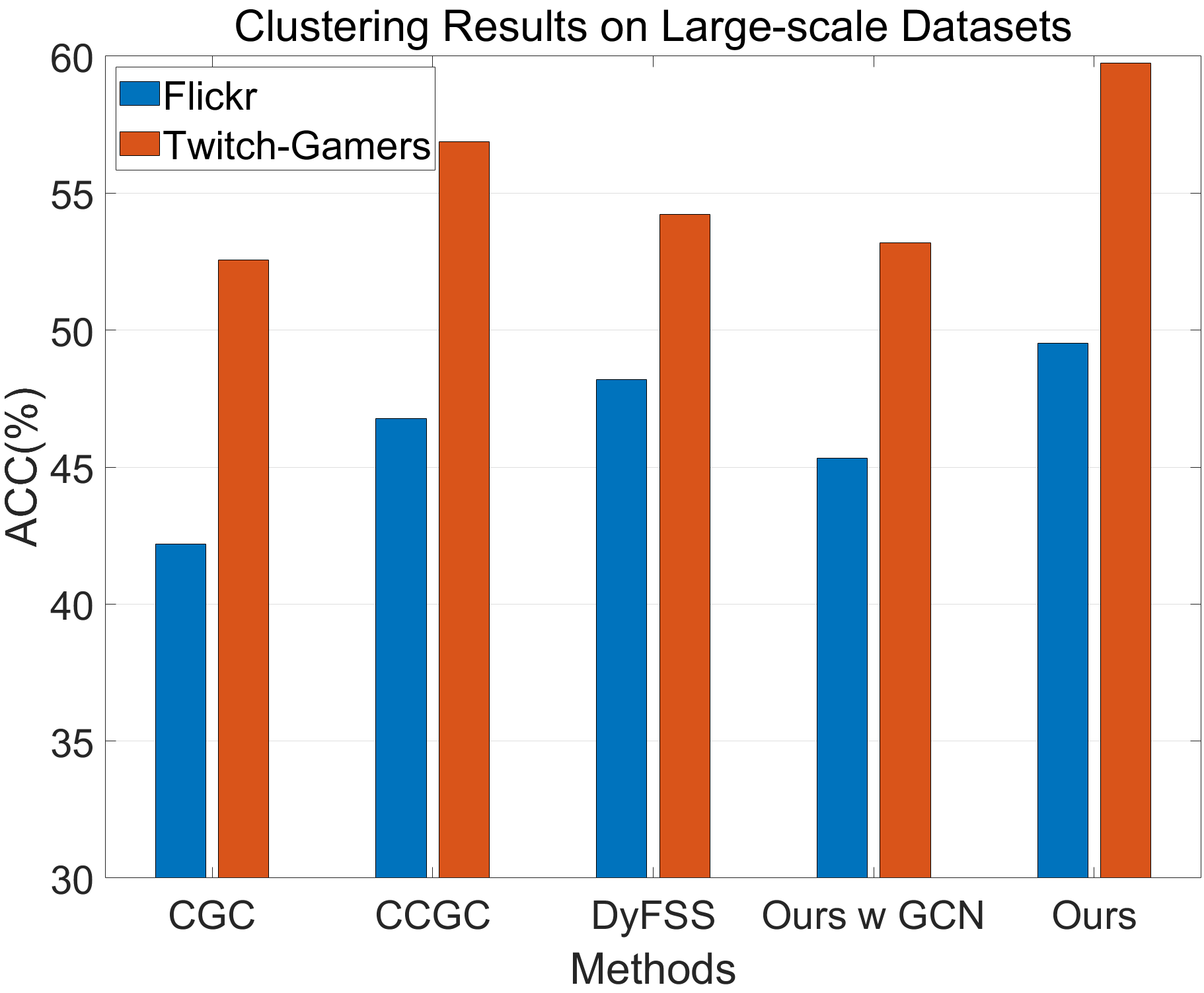}
\caption{Clustering results on large-scale graphs.}
\label{larged}
\end{figure}

\begin{table*}[htbp]
		\centering
  \caption{Clustering results on heterophilic graphs. The best results are marked in \textcolor[rgb]{ 1,  0,  0}{\textbf{red}}. The runner-up is marked in \textbf{bold}.}
		\label{Reheter}%
    \resizebox{.9\textwidth}{!}{
    \begin{tabular}{ccccccccccccc}
    \toprule
    \multirow{2}[4]{*}{Methods} & \multicolumn{2}{c}{Cornell} & \multicolumn{2}{c}{Wisconsin} & \multicolumn{2}{c}{Washington} & \multicolumn{2}{c}{Chameleon} & \multicolumn{2}{c}{Squirrel}  & \multicolumn{2}{c}{Roman-empire}\\
\cmidrule{2-13}          & ACC   & NMI   & ACC   & NMI   & ACC   & NMI   & ACC   & NMI   & ACC   & NMI   & ACC   & NMI \\
    \midrule
    DAEGC & 42.56  & 12.37 &39.62 &12.02 & 46.96  &17.03   & 32.06  & 6.45   & 25.55  & 2.36 &21.23 &12.67\\
    MSGA  & 50.77  & 14.05 &54.72 &16.28 & 49.38   & 6.38   & 27.98   &6.21   & 27.42  & 4.31 &19.31 &12.25\\
    FGC   & 44.10  & 8.60  &50.19 &12.92 & 57.39  & 21.38   & 36.50 &11.25   & 25.11  & 1.32 &14.46 &4.86 \\
    GMM  & 58.86 & 9.26   &51.70 &9.68  & 60.86  & 20.56 & 34.91   & 7.89   &29.76   &5.15 & 21.90   & 13.57 \\
    RWR  & 58.29 & 11.35 &53.58 &16.25 & 63.91  & 23.13 & 33.27   & 8.03   & 29.96   & 5.69 & 22.73   & 14.74\\
    ARVGA  & 56.23  & 9.35   &54.34 &11.41 & 60.87  & 16.19 & 37.33   & 9.77   & 25.32   & 2.88 & 22.89   & 15.25\\
    DCRN &51.32 &9.05 &57.74 &19.86 &55.65 &14.15 &34.52 &9.11 &30.69 &6.84 &32.57 &29.50 \\
    SELENE &57.96&17.32 &65.23&25.40 &-&- &\textbf{38.97}&\textbf{20.63} &-&- &-&-\\
    CGC &44.62 &14.11 &55.85 &23.03 &63.20&25.94 &36.43&11.59 &27.23 &2.98 &30.16 &27.25\\
    DGCN & \textbf{62.29}  &\textbf{29.93} &\textbf{71.71} &\textcolor[rgb]{ 1,  0,  0}{\textbf{41.29}} & \textbf{69.13} &28.22 &36.14  &11.23 &\textbf{31.34} &7.24 & 33.42   &\textbf{31.44} \\
    RGSL &57.44 &28.95 &56.60 &28.57 &66.09 &\textbf{29.79} &38.52 &12.79 &30.74 &\textbf{8.74} &\textbf{34.57} &31.23 \\

    AGCN &\textcolor[rgb]{ 1,  0,  0}{\textbf{68.31}}&\textcolor[rgb]{ 1,  0,  0}{\textbf{41.35}} &\textcolor[rgb]{ 1,  0,  0}{\textbf{72.11}}&\textbf{37.54} &\textcolor[rgb]{ 1,  0,  0}{\textbf{70.43}} &\textcolor[rgb]{ 1,  0,  0}{\textbf{34.88}} &\textcolor[rgb]{ 1,  0,  0}{\textbf{41.06}}&\textcolor[rgb]{ 1,  0,  0}{\textbf{21.07}}&\textcolor[rgb]{ 1,  0,  0}{\textbf{34.07}}&\textcolor[rgb]{ 1,  0,  0}{\textbf{8.94}} &\textcolor[rgb]{ 1,  0,  0}{\textbf{38.42}}&\textcolor[rgb]{ 1,  0,  0}{\textbf{37.26}}\\

    \bottomrule
    \end{tabular}}%
	\end{table*}%

 \begin{table*}[!htbp]
		\centering
  \caption{Clustering results on homophilic graphs.}
  % \textcolor{red}{where is the rank?}
		\label{Rehomo}%
    \resizebox{.6\textwidth}{!}{
        \begin{tabular}{cccccccccc}
    \toprule
    \multirow{2}[4]{*}{Methods} & \multicolumn{2}{c}{Cora} & \multicolumn{2}{c}{Citeseer}  & \multicolumn{2}{c}{Pubmed} & \multicolumn{2}{c}{UAT}  \\%& \multirow{2}[4]{*}{AvgRank} \\
\cmidrule{2-9}          & ACC   & NMI   & ACC   & NMI   & ACC   & NMI   & {ACC} & NMI   &  \\
    \midrule
    DFCN  & 36.33  & 19.36 & 69.50  & 43.9  &-&-& 33.61  & 26.49   \\
    %DCRN  & 48.93  & -   & 70.86  & \textcolor[rgb]{ 1,  0,  0}{\textbf{45.86}} &-&-& -  &- & {\textcolor[rgb]{ 1,  0,  0}{\textbf{79.94}}} & \textcolor[rgb]{ 1,  0,  0}{\textbf{73.70 }}    \\
    SSGC  & 69.60  & 54.71 & 69.11  & 42.87 &-&- & 36.74  & 8.04   \\
    MVGRL & 70.47  & 55.57 & 68.66  & 43.66 &-&-& 44.16  & 21.53   \\
    SDCN  & 60.24  & 50.04 & 65.96  & 38.71 &65.78&29.47& 52.25  & 21.61   \\
    MCGC  & 42.85  & 24.11 & 64.76  & 39.11 &66.95&32.45& 41.93  & 16.64   \\
    FGC   & 72.90  & 56.12 & 69.01  & 44.02&70.01&31.56 & 53.03  & 27.06   \\
    SCGC & 73.88 & 56.10 &71.02 &\textbf{45.25} &67.73&28.65&\textbf{56.58} &28.07 \\
    CCGC & 73.88 & 56.45 & 69.84 & 44.33 &68.06&30.92&56.34 &\textbf{28.15}   \\
    CGC & \textbf{75.15} & \textbf{56.90} &69.31 &43.61 &67.43&33.07&49.58&17.49 \\
    DGCN & 72.19 & 56.04 & \textbf{71.27} & 44.13 &70.13&30.17& 52.27 & 23.54 \\
    %DMGNC & 73.12 & 54.80 & \textbf{71.27} & 44.40 &\textbf{70.46}&\textbf{34.21}&- &- \\
    DyFSS &72.19 &55.49 &70.18 &44.80 &68.05 &26.87 &51.43 &25.52 \\
    GTAGC &71.70 &54.00 &70.80	&45.20 &67.80	&31.80  &-  &- \\
    TGAE &72.60	&52.90 &69.60	&43.00 &\textcolor[rgb]{ 1,  0,  0}{\textbf{76.30}}	&\textcolor[rgb]{ 1,  0,  0}{\textbf{37.00}} &-&- \\
    AGCN & \textcolor[rgb]{ 1,  0,  0}{\textbf{75.95}} & \textcolor[rgb]{ 1,  0,  0}{\textbf{57.25}} & \textcolor[rgb]{ 1,  0,  0}{\textbf{71.93}} &\textcolor[rgb]{ 1,  0,  0}{\textbf{45.81}} &\textbf{72.58} &\textbf{35.43} & \textcolor[rgb]{ 1,  0,  0}{\textbf{60.84}} &\textcolor[rgb]{ 1,  0,  0}{\textbf{30.74}}  \\
    \bottomrule
    \end{tabular}}%
	\end{table*}%

\subsection{Clustering Results}
We begin by experimenting on heterophilic datasets, where long-range information is advantageous. AGCN achieves state-of-the-art results on all datasets in Table \ref{Reheter}. AGCN clearly outperforms recent baselines that handle heterophily, such as SELENE, CGC, DGCN, and RGSL. Specifically, AGCN outperforms these methods by margins of up to 6.02\% and 11.42\% on the Cornell dataset, and 3.85\% and 5.82\% on the Roman-Empire dataset. Note that DGCN, CGC, and RGSL model graph structures using complex optimization equations or the contrastive learning approach. Thus, modeling graph structures through the attention mechanism shows promise. The remaining methods are based on traditional GNNs. AGCN improves ACC and NMI by at least 9.45\% and 28.98\% on the Cornell dataset, and 5.85\% and 7.76\% on the Roman-Empire dataset, further highlighting the effectiveness of the attention mechanism in handling graph structures compared to traditional GNNs.

We also evaluate our approach on the homophilic datasets in Table \ref{Rehomo}. AGCN achieves the best performance on 3 out of 4 datasets, and the second-best performance on the remaining dataset. GTAGC and TGAE are the latest Transformer-based graph clustering methods, which apply the vanilla Transformer directly. The high performance of AGCN, especially on the Cora dataset, highlights the importance of making the Transformer structure-aware. AGCN performs worse than TGAE on Pubmed, which may be due to the fact that Pubmed contains only three clusters. The limited number of clusters increases the likelihood that the Transformer’s attention mechanism focuses on nodes within the same cluster. Compared to other GNN-based methods, GTAGC and TGAE show poor performance, indicating that the vanilla Transformer is not favored in graph clustering. However, AGCN outperforms all of them in every case. Therefore, the attention mechanism can also be effective for modeling homophilic graphs in clustering, potentially outperforming traditional GNNs.

To evaluate AGCN's scalability, we further test it on two large datasets, Flickr and Twitch-Gamer, which contain 89,250 and 168,114 nodes, respecti
vely. Most baselines from Tables \ref{Reheter} and \ref{Rehomo} fail to perform on these datasets due to their high complexity, such as DGCN. We evaluate three recent baselines and also replace our attention network with GCN, denoted as ``Ours w GCN''. The results are presented in Fig. \ref{larged}. AGCN demonstrates the best performance on these datasets, while AGCN with GCN performs poorly. This further emphasizes its effectiveness in scaling to large datasets, which is attributed to the KV Cache idea.

In summary, AGCN remains effective even in the absence of homophily, a common scenario in real-world applications. Moreover, AGCN is scalable, making it suitable for large-scale deployment.

\subsection{Benefits of Capturing Long-range Information in Graph Clustering}

\begin{table}
\begin{minipage}{1\linewidth}
\centering
\caption{The results with the masked features. The best performance is marked in \textbf{bold}.}\label{lrange}
\begin{center}
\resizebox{1.\textwidth}{!}{
\begin{tabular}{l rl| rl| rl| rl|}
\midrule
Methods & \multicolumn{2}{c}{{CCGC}} & \multicolumn{2}{c}{{AGCN w GCN}} & \multicolumn{2}{c}{{AMGC}} & \multicolumn{2}{c}{{AGCN}} \\ 
\cmidrule(lr){2-3} \cmidrule(lr){4-5} \cmidrule(lr){6-7} \cmidrule(lr){8-9}
    & {ACC} & {NMI} & {ACC} & {NMI}& {ACC} & {NMI}& {ACC} & {NMI}\\
\midrule
Cora
&37.93 &22.37
&55.20 &56.32
&66.65 &47.99
&\textbf{68.76} &\textbf{49.22}
\\
Citeseer
&39.63 &17.13
&37.68 &13.44
&60.92 &32.93
&\textbf{61.23} &\textbf{33.21}
 \\
Pubmed
&42.70 &2.43
&41.56&15.23
&64.56 &24.58
&\textbf{65.14} &\textbf{25.17}
\\
\bottomrule
\end{tabular}}
\end{center}
\end{minipage}
\end{table}

%\subsubsection{Sparse feature setting}
%To empirically validate the effectiveness of incorporating the global information in an unsupervised setting, we propose a feature-masking strategy: we randomly mask part of the features and evaluate whether the model can still recover cluster-consistent representations. This simulates scenarios where critical information is partially missing and emphasizes the need for efficient information propagation across the graph.
To empirically assess the impact of capturing long-range information, we examine a generalized clustering scenario with sparsely distributed features, where critical discriminative information must propagate across large gaps to reach the majority of feature-missing data. In particular, AMGC \cite{tu2024attribute} is the most recent work addressing the problem of feature-missing data in graph clustering. Therefore, we replicate its experiments for sparse feature settings by removing all features for 60\% of the nodes at random. We also replace our structure-aware Transformer with GCN, denoted as ``AGCN w GCN''.

We hypothesize that bridging long-range dependencies becomes more crucial when local node information is insufficient, making global information especially beneficial. This hypothesis is supported by the experimental results in Table \ref{lrange}. It can be seen that traditional GNN-based clustering methods CCGC and ``AGCN w GCN'' achieve poor performance, thus traditional GNNs fail with sparse features. Besides, AGCN continues to achieve the best performance compared to AMGC.

\begin{figure}[!htbp]
    %\vspace{-18pt}
    %\hspace*{5em}
    \centering
    \includegraphics[width=.8\linewidth]{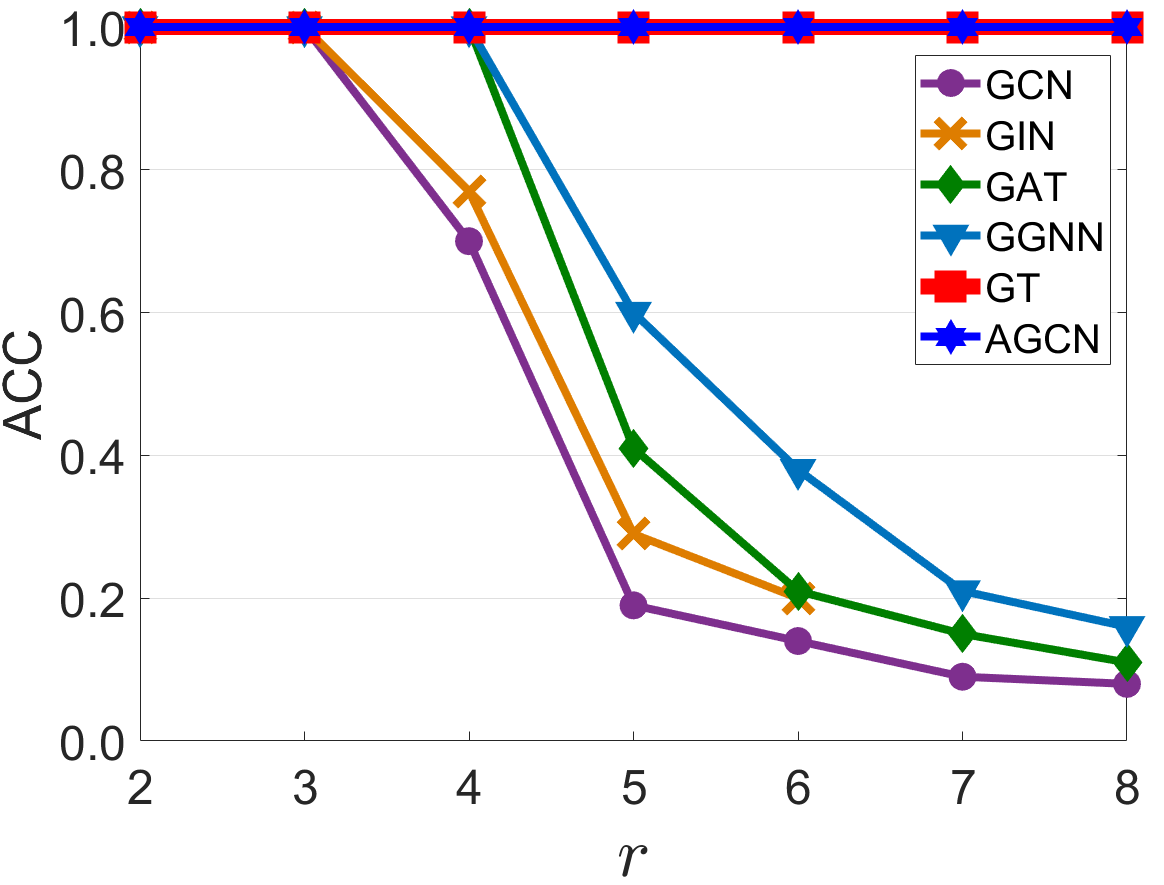}
    \caption{Experimental results on the Tree-Neighbors Match dataset.}
    \label{LRR}
\end{figure}

Further, following the protocol in GT \cite{mullerattending}, we conduct experiments on the Tree-Neighbors Match dataset. This benchmark specifically requires models to capture long-range interactions between leaf nodes and the root across varying tree depths. The baselines and the results are directly quoted from the original paper. Let $r$ denote the tree depth, and we report accuracy results accordingly in Fig. \ref{LRR}. Our method, AGCN, achieves perfect performance—matching GT—across all depth settings and significantly outperforming standard message-passing GNNs. These results demonstrate AGCN's strong capability in modeling long-range dependencies.

%we randomly mask the features on Cora with the ratio $Nr$. We select two recent baselines with GNNs, and replace our structure-aware Transformer with GCN, which is marked as ``Ours w GCN''. As shown in Fig. \ref{lrange} (a), our method achieves the best performance and shows a more stable trend compared to the other baselines. Therefore, AGCN can capture global information and is robust to the noise.

\begin{table}
\vspace{-5pt}
\begin{minipage}{1\linewidth}
\centering
\caption{Results of ablation study. The best performance is marked in \textbf{bold}.}\label{abla1}
\begin{center}
\resizebox{1.\textwidth}{!}{
\begin{tabular}{l rl| rl| rl}
\midrule
Methods & \multicolumn{2}{c}{{AGCN w/o $L_{neg}$}} & \multicolumn{2}{c}{{AGCN w $T$}} & \multicolumn{2}{c}{{AGCN}} \\ 
\cmidrule(lr){2-3} \cmidrule(lr){4-5} \cmidrule(lr){6-7} 
    & {ACC} & {NMI} & {ACC} & {NMI}& {ACC} & {NMI}\\
\midrule
Cora
&74.81 &56.74
&71.60 &52.47
&\textbf{75.95} &\textbf{57.25}
\\
UAT
&58.43 &28.76
&52.54 &23.88
&\textbf{60.84} &\textbf{30.74}
 \\
Cornell
&65.57 &34.62
&56.46 &34.39
&\textbf{68.31} &\textbf{41.35}
\\
Washington
&63.42 &27.76
&65.71 &28.72
&\textbf{70.43} &\textbf{34.88}
\\
\bottomrule
\end{tabular}}
\end{center}
\end{minipage}
\end{table}

\subsection{Ablation Study}
We conduct ablation studies to explore the functionality of AGCN's components, i.e., the proposed structure-aware Transformer and the pair margin loss. Table \ref{abla1} shows the results on Cora, UAT, Cornell, and Washington. ``AGCN w/o $L_{neg}$'' indicates the results without our proposed $L_{neg}$, i.e., positive pairs are directly selected by the connected nodes. It can be seen that the performance degrades in all cases, thus, our proposed contrastive loss can better characterize the relationship between the connected nodes. We also replace our structure-aware Transformer with vanilla Transformer, i.e., \(H^{(l+1)} = softmax(\frac{(H^{(l)}W^{(l+1)}_Q)(H^{(l)}W^{(l+1)}_K)^\top}{\sqrt {d_K}}) H^{(l)}W_V^{(l+1)}\), which is marked as ``AGCN w $T$''. The performance decreases dramatically in all cases. Therefore, alleviating the problem of over-globalization through structure awareness is crucial in clustering.

We also conduct experiments to evaluate the efficiency of our method. Using the Cora and Roman-empire datasets, we compare AGCN with state-of-the-art methods in running time. The results are shown in Table \ref{time}. It can be seen that AGCN is fast. This is because our proposed structure-aware Transformer only requires linear complexity and the contrastive learning loss is designed to be minimalist. Therefore, AGCN incurs no additional runtime compared to our baselines.

\begin{table}[ht]
\centering
\caption{Analysis of total running time (seconds) on Cora and Roman-Empire dataset. The best performance is marked in bold.}
\label{tab:runtime}
\begin{tabular}{lcccc}
\hline
Methods & RGSL & DGCN & DyFSS & AGCN \\
\hline
Cora (s) & 125.43 & 97.42 & 55.33 & \textbf{35.45} \\
Roman-Empire (s) & 2721.32 & 1862.14 & 1412.73 & \textbf{1259.89} \\
\hline
\end{tabular}\label{time}
\end{table}

\subsection{Sensitivity Analysis}
\begin{figure}[!htbp]
    %\vspace{-18pt}
    %\hspace*{5em}
    \centering
    \includegraphics[width=1.\linewidth]{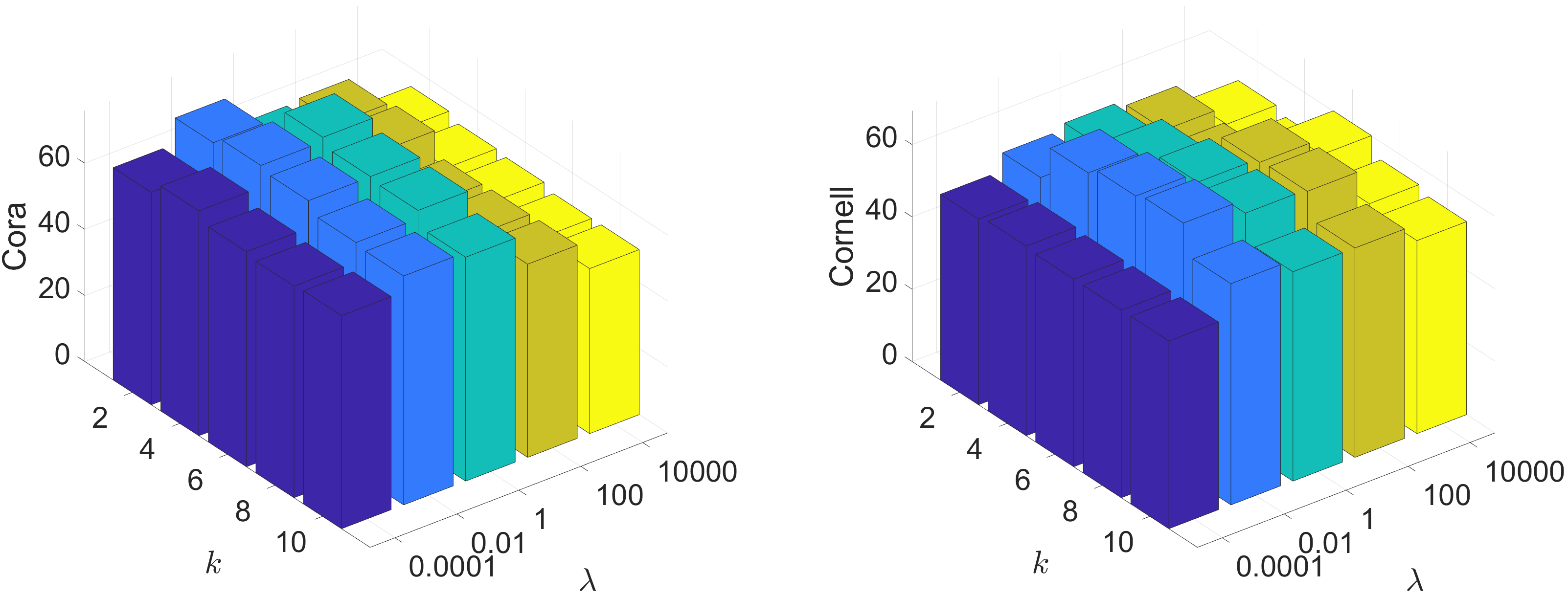}
    \caption{Sensitivity analysis of $k$ and $\lambda$ on Cora (left) and Cornell (right).}
    \label{sens}
\end{figure}

Our model has two parameters: $k$, which controls global information, and $\lambda$, a balance parameter. To observe their impact, we vary $k$ within the range \{2, 4, 6, 8, 10\} and $\lambda$ within the range \{1e-4, 1e-2, 1, 1e2, 1e4\}. Fig. \ref{sens} shows that our model works well across a large scale, though extremely large or small values tend to result in poor outcomes. This stability highlights the model’s ability to generalize across varying conditions. Additionally, Cora prefers a relatively small $k$, while Cornell prefers a larger $k$. This is because homophilic graphs capture more information from low-order neighbors, while the receptive field needs to be enlarged in heterophilic graphs. Additionally, both Cora and Cornell prefer a middle value $\lambda$, suggesting that a balance between positive and negative samples yields better results.

\section{Conclusion}
In this work, we revisit the role of attention in graph structures and propose AGCN. We identify that traditional GNNs tend to overemphasize the grouping effect, while Transformers suffer from overly global attention in unsupervised scenarios. Our method addresses both issues by directly modeling the graph structure with the attention mechanism and introducing a pairwise contrastive loss to enhance the discriminative power of the representations. Additionally, our design benefits from improved scalability due to the KV cache mechanism. Extensive experiments on various homophilic and heterophilic graph datasets demonstrate that AGCN consistently outperforms current baselines in clustering performance.

\bibliography{1}
\bibliographystyle{IEEEtran}

%\vfill

\end{document}